\documentclass[11pt,a4paper]{article}
\usepackage[hyperref]{emnlp2020}
\usepackage{times}
\usepackage{latexsym}
\usepackage{graphicx}
\usepackage{subcaption}
\usepackage{enumitem}
\usepackage{multicol}
\usepackage{multirow}
\usepackage{makecell}
\usepackage{amsmath}
\usepackage{algorithm}
\usepackage{algorithmic}
\usepackage{enumitem}

\usepackage{microtype}
\definecolor{applegreen}{rgb}{0.55, 0.71, 0.0}

\aclfinalcopy 
\def\aclpaperid{3529} 

\title{Towards Interpreting BERT for Reading Comprehension Based QA}

\author{Sahana Ramnath\thanks{* now at Google Research, Bangalore}, Preksha Nema, Deep Sahni, Mitesh M. Khapra  \\
Robert Bosch Centre for Data Science and AI (RBC-DSAI) \\ Indian Institute of Technology Madras, Chennai, India \\ 
\texttt{\{sahanjich,deep.sahani11\}@gmail.com,} \\ \texttt{\{preksha,miteshk\}@cse.iitm.ac.in}}

\date{}

\begin{document}
\maketitle
\begin{abstract}
BERT and its variants have achieved state-of-the-art performance in various NLP tasks. Since then, various works have been proposed to analyze the linguistic information being captured in BERT. However, the current works do not provide an insight into how BERT is able to achieve near human-level performance on the task of Reading Comprehension based Question Answering. In this work, we attempt to interpret BERT for RCQA. Since BERT layers do not have predefined roles, we define a layer's role or functionality using Integrated Gradients. Based on the defined roles, we perform a preliminary analysis across all layers. We observed that the initial layers focus on query-passage interaction, whereas later layers focus more on contextual understanding and enhancing the answer prediction. Specifically for quantifier questions (how much/how many), we notice that BERT focuses on confusing words (i.e., on other numerical quantities in the passage) in the later layers, but still manages to predict the answer correctly. The fine-tuning and analysis scripts will be publicly available at \url{https://github.com/iitmnlp/BERT-Analysis-RCQA}.
\end{abstract}

\section{Introduction}
The past decade has witnessed a surge in the development of deep neural network models to solve NLP tasks. Pretrained language models such as ELMO \cite{elmo}, BERT \cite{Devlin:18} , XLNet \cite{xlnet} etc. have achieved state-of-the-art results on various NLP tasks. This success motivated various studies to understand how BERT achieves human-level performance on these tasks. ~\citet{tenney2019bert,peters2018dissecting} analyze syntactic and semantic roles played by different layers in such models. \citet{Clark:19} specifically analyze BERT's attention heads for syntactic and linguistic phenomena. Most of these works focus on tasks such as sentiment classification, syntactic/semantic tags prediction, natural language inference, and so on.
However, to the best of our knowledge, BERT has not been thoroughly analyzed for complex tasks like RCQA. It is a challenging task because of 1) the large number of parameters and non-linearities in BERT, and 2) the absence of pre-defined roles across layers in BERT as compared to pre-BERT models like BiDAF \citep{Seo:16} or DCN \citep{Xiong:16}. In this work, we take the first step to identify each layer's role using the attribution method of Integrated Gradients \cite{Sundararajan:17}. We then try to map these roles to the following functions, deemed necessary in pre-BERT models to reach the answer: (i) learn contextual representations for the passage and the question, individually, (ii) attend to information in the passage specific to the question and, (iii) predict the answer. 

\noindent We perform analysis on the SQuAD \citep{Rajpurkar:16} and DuoRC \citep{Saha:18} datasets. We observe that the initial layers primarily focus on question words that are present in the passage. In the later layers, the focus on question words decreases, and more focus is on the supporting words that surround the answer and the predicted answer span. Further, through a focused analysis of quantifier questions (questions that require a numerical entity as the answer), we observe that BERT pays importance to many words similar to the answer (same type, such as numbers) in later layers. We find this intriguing since, even after marking confusing words spread across passage as important, BERT's prediction accuracy is high. We also provide qualitative analysis to demonstrate the above trends. 

\section{Related Work}
In the past few years, various large-scale datasets have been proposed for the RCQA task ~\citep{Nguyen:16,Joshi:17, Rajpurkar:16,Saha:18} which have led to various deep neural-network (NN) based architectures such as ~\citet{Seo:16, dhingra2016gated}. Additionally, with complex pretraining, models such as ~\citet{Liu:19,Lan:19,Devlin:18} are very close to human-level performance. 
Due to the large number of parameters and non-linearity of deep NN models, the answer to the question ``how did the model arrive at the prediction?'', is not known; hence, they are termed as \textit{blackbox models}. Motivated by this question, there have also been many works that analyze the interpretability of deep NN models on NLP tasks; many of them analyze models based on in-built attention mechanisms \citep{Jain:19, serrano2019attention, wiegreffe2019attention}. Further, various attribution methods such as ~\citet{Bach:15, Sundararajan:17} have been proposed to analyze them. \citet{tenney2019bert} and \citet{peters2018dissecting} perform a layerwise analysis of BERT and BERT-like models to assign them syntactic and semantic meaning using probing classifiers. \citet{si2019does} question BERT's working on QA tasks through adversarial attacks, similar to \citet{Jia:17, Mudrakarta:18}. They point out that BERT is prone to be fooled by such attacks. Unlike these earlier works, we focus on analyzing BERT's layers specifically for RCQA to understand their QA-specific roles and their behavior on potentially confusing quantifier questions.
\section{Experimental Setup}
For our BERT analysis, we use the BERT-BASE model, which has 12 Transformer blocks (layers), each with a multi-head self-attention and a feed-forward neural network. We use the official code and pre-trained checkpoints\footnote{\url{https://github.com/google-research/bert}} and fine-tune it for two epochs for the SQuAD and DuoRC datasets to achieve F1 scores of $88.73$ and $54.80$ on their respective dev-splits. We use SQuAD \cite{Rajpurkar:16} 1.1 with 90k/10k train/dev samples, each with a 100-300 words passage and the SelfRC dataset in DuoRC \cite{Saha:18} with 60k/13k train/dev samples, each with a 500 (on average) words passage. For each passage, both datasets have a natural language query and answer span in the passage itself. 
\section{Layer-wise Functionality} \label{layerwise_ig}
As discussed earlier, we aim to understand each BERT layer's functionality for the RCQA task; we want to identify the passage words that are of primary importance at each layer for the answer. Intuitively, the initial layers should focus on question words, and the latter should zoom in on contextual words that point to the answer. To analyze the above, we use the attribution method Integrated Gradients \citep{Sundararajan:17} on BERT at a layerwise level. 

For a given passage P consisting of $n$ words $[w_1,w_2,\dots,w_n]$, query $Q$, and model $f$ with $\theta$ parameters, answer prediction is modeled as:
\begin{align}
    \nonumber p(w_s, w_e) &= f(w_s, w_e|P,Q,\theta)
\end{align}
where $w_s,w_e$ are the predicted answer start and end words or positions.

\noindent For any given layer \textit{l}, the above is equivalent to:
\begin{align}
    \nonumber p(w_s, w_e) &= f_l(w_s, w_e|E_{l-1}(P),E_{l-1}(Q),\theta)
\end{align}
where $f_l$ is the forward propagation from layer $l$ to the prediction. $E_{l}(.)$, is the representation learnt for passage or query words by a given layer \textit{l}. To elaborate, we consider the network below the layer \textit{l} as a blackbox which generates \textit{input} representations for layer \textit{l}. 
The \textbf{Integrated Gradients} for a Model $M$, a passage word $w_i$, embedded as $x_i \in \mathbf{R}^L$ is:
\begin{align}
\nonumber IG(x_i) =  \int\limits_{\alpha=0}^1\frac{\partial M(\tilde{x} + \alpha(x_i - \tilde{x}))}{\partial x_i}\  \ d\alpha  
\end{align}
where $\tilde{x}$ is a zero vector, that serves as a baseline to measure integrated gradient for $w_i$. 
 We calculate the integrated gradients at each layer $IG_l(x_i)$ for all passage words $w_i$ using Algorithm \ref{imp_distribution}. We approximate the above integral across $50$ uniform samples of $\alpha \in [0,1]$. We then compute importance scores for each $w_i$ by taking the euclidean norm of $IG(w_i)$ and normalizing it to get a probability distribution $I_l$ over the passage words.  
\begin{algorithm}
\caption{To compute Layer-wise Integrated Gradients for layer \textit{l}}
\label{imp_distribution}
\begin{algorithmic}[1]
\STATE $\tilde{p} = 0$  \quad //zero baseline \\
\STATE $m=50$ \\
\STATE $G_l(p) = \frac{1}{m}\sum_{k=1}^{m}\frac{\partial f_l(\tilde{p} + \frac{k}{m}(p - \tilde{p}))}{\partial E_l}$
\STATE $IG_l(p) = [(p-\tilde{p}) \times G_l(p)] $
\STATE // Compute squared norm for each word \\
\STATE $\tilde{I}_l([w_1,\dots,w_k]) = ||IG_l(p)|| \in \mathrm{R}^k$
\STATE Normalize $\tilde{I}_l$ to a probability distribution $I_l$
\end{algorithmic}
\end{algorithm}
\subsection{JSD with top-k retained/removed} \label{subsec:topk}
\begin{figure}
    \centering
    \begin{subfigure}{0.22\textwidth}
    \centering
    \includegraphics[width=\textwidth]{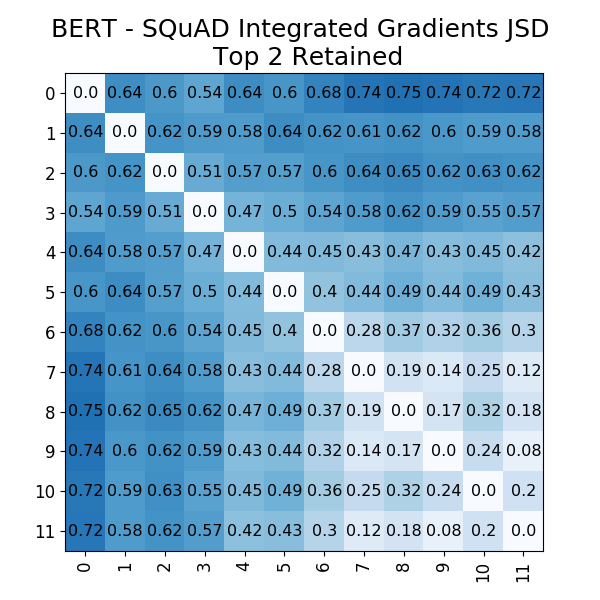}
    \caption{}
    \label{bert_jsd1}
    \end{subfigure}
    \begin{subfigure}{0.22\textwidth}
    \centering
    \includegraphics[width=\textwidth]{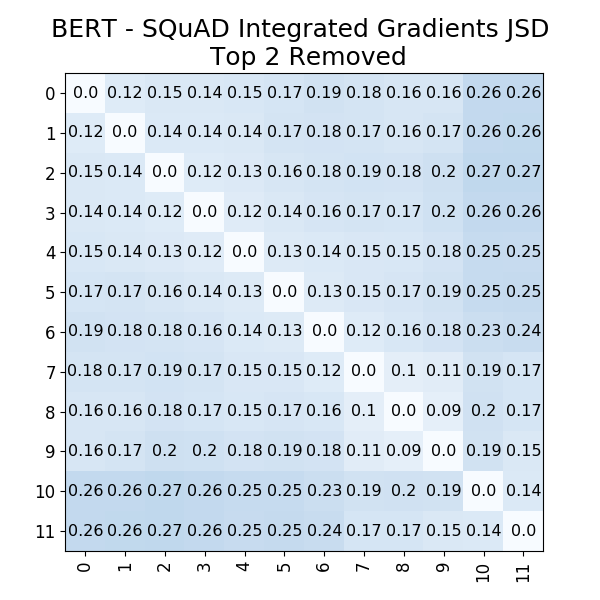}
    \caption{}
    \label{bert_jsd2}
    \end{subfigure} \\
    \begin{subfigure}{0.22\textwidth}
    \centering
    \includegraphics[width=\textwidth]{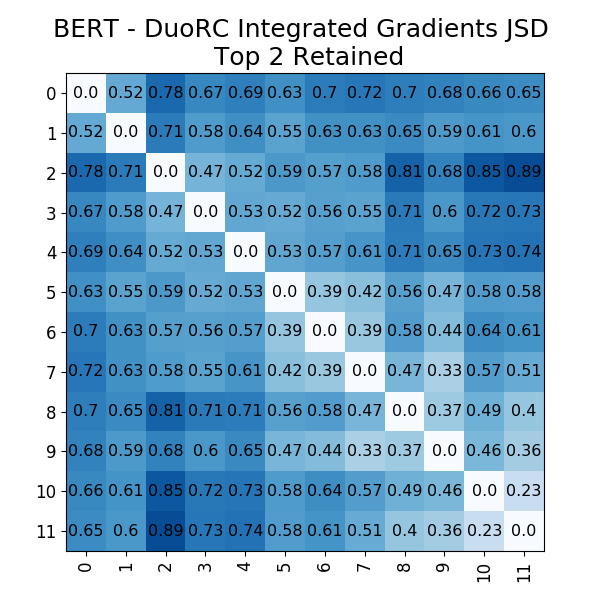}
    \caption{}
    \label{bert_jsd3}
    \end{subfigure}
    \begin{subfigure}{0.22\textwidth}
    \centering
    \includegraphics[width=\textwidth]{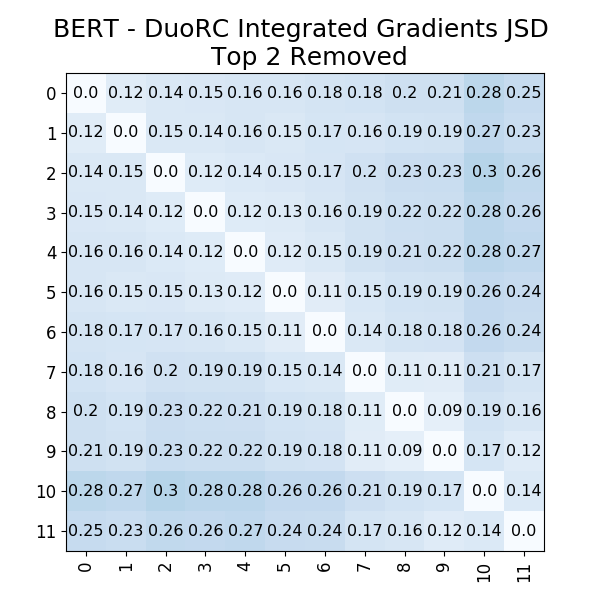}
    \caption{}
    \label{bert_jsd4}
    \end{subfigure}    
    \caption{JSD between $I_l$'s with top-2 items removed/retained (SQuAD - (a), (b), DuoRC - (c), (d))}
    \label{fig:bert_jsd_rem_keep}
\end{figure}

We quantify and visualize a layer's function as its distribution of importance over the passage words $I_l$. To compute the similarity between any two layers $x,y$, we measure the \textit{Jensen-Shannon Divergence (JSD)} between their corresponding importance distributions $I_x, I_y$. We calculate the JSD scores between every pair of layers in the model and visualize it as a $n_l \times n_l$ heatmap ($n_l$ - number of layers in the model). A higher JSD score corresponds to the two layers being more different. This further means the two layers consider different words as salient. We visualize heatmaps for the dev-splits of SQuAD (Figures \ref{bert_jsd1}, \ref{bert_jsd2}) and DuoRC (Figures \ref{bert_jsd3}, \ref{bert_jsd4}), averaging over 1000 samples in each case.\\

We analyze the distribution in two parts: (i) we retain only top-k scores in each layer and zero out the rest, which denotes the distribution's head. (ii) we zero the top-k scores in each layer and retain the rest, which denotes the distribution's tail.  In either case, we re-normalize to get a probability distribution.
When comparing just the top-2 items, we see higher values (min 0.08/max 0.72) in heatmap \ref{bert_jsd1} than in heatmap \ref{bert_jsd2} (min 0.09/max 0.26). Similarly, we see higher values (min 0.23/max 0.89) in heatmap \ref{bert_jsd3} than in heatmap \ref{bert_jsd4} (min 0.12/max 0.28).We conclude that a layer's function is reflected in words high up in the importance distribution. As we remove them, we encounter an almost uniform distribution across the less important words. Hence to correctly identify a layer's functionality, we need to focus only on the head (top-k words) and not on the tail.
\section{Results and Discussions}

\begin{table}
\centering
\resizebox{0.85\linewidth}{!}{%
\begin{tabular}{|c|c|c|c|}
\hline
\textbf{Layer Name} & \textbf{\begin{tabular}[c]{@{}c@{}}\% answer\\ span\end{tabular}} & \textbf{\% Q-words} & \textbf{\begin{tabular}[c]{@{}c@{}}\% Contextual\\ Words\end{tabular}} \\ \hline \hline
Layer 0 & 26.99 & 22.94 & 9.45 \\ \hline
Layer 1 & 26.09 & 24.35 & 9.43 \\ \hline
Layer 2 & 29.9 & 22.41 & 11.65 \\ \hline
Layer 3 & 30.44 & 19.55 & 11.13 \\ \hline
Layer 4 & 30.06 & 18.33 & 11.23 \\ \hline
Layer 5 & 30.75 & 14.71 & 11.57 \\ \hline
Layer 6 & 31.25 & 15.33 & 11.94 \\ \hline
Layer 7 & 32.37 & 12.29 & 12.32 \\ \hline
Layer 8 & 30.78 & 18.91 & 12.07 \\ \hline
Layer 9 & 34.58 & 10.21 & 13.41 \\ \hline
Layer 10 & 34.31 & 10.56 & 13.39 \\ \hline
Layer 11 & 34.63 & 12.0 & 13.74 \\ \hline
\end{tabular}%
}
\caption{Semantic statistics of top-5 words - SQuAD}
\label{tab:semantic_stats_squad}
\end{table}

\begin{table}
\centering
\resizebox{0.85\linewidth}{!}{%
\begin{tabular}{|c|c|c|c|}
\hline
\textbf{Layer Name} & \textbf{\begin{tabular}[c]{@{}c@{}}\% answer\\ span\end{tabular}} & \textbf{\% Q-words} & \textbf{\begin{tabular}[c]{@{}c@{}}\% Contextual\\ Words\end{tabular}} \\ \hline \hline
Layer 0 & 35.14 & 17.89 & 27.53 \\ \hline
Layer 1 & 37.29 & 18.29 & 29.88 \\ \hline
Layer 2 & 38.30 & 19.59 & 30.05 \\ \hline
Layer 3 & 34.37 & 18.88 & 25.83 \\ \hline
Layer 4 & 33.93 & 20.77 & 26.20 \\ \hline
Layer 5 & 36.32 & 16.16 & 27.97 \\ \hline
Layer 6 & 35.34 & 15.75 & 27.05 \\ \hline
Layer 7 & 41.20 & 10.57 & 31.12 \\ \hline
Layer 8 & 40.38 & 8.50 & 22.16 \\ \hline
Layer 9 & 41.25 & 8.03 & 17.9 \\ \hline
Layer 10 & 43.93 & 5.58 & 15.85 \\ \hline
Layer 11 & 44.37 & 6.00 & 33.74 \\ \hline
\end{tabular}%
}
\caption{Semantic statistics of top-5 words - DuoRC}
\label{tab:semantic_stats_duo}
\end{table}
\begin{table*}
    \small
    \begin{tabular}{p{15cm}}
        \Xhline{3\arrayrulewidth}
        \textbf{Question:}Why was Polonia relegated from the country's top flight in 2013? \\
        \textbf{Answer:} disastrous financial situation\\
         \hline 
             \begin{tabular}{p{0.25cm} p{6.5cm} p{0.25cm} p{6.5cm}}
                \textbf{L0} & {\setlength{\fboxsep}{0pt}\colorbox[Hsb]{202, 0.35, 1.0}{\strut Polonia}} was {\setlength{\fboxsep}{0pt}\colorbox[Hsb]{202, 0.16, 1.0}{\strut relegated}} from the country's top flight in 2013 because of their {\setlength{\fboxsep}{0pt}\colorbox[Hsb]{202, 0.25, 1.0}{\strut disastrous}} financial {\setlength{\fboxsep}{0pt}\colorbox[Hsb]{202, 0.5, 1.0}{\strut situation}}{\setlength{\fboxsep}{0pt}\colorbox[Hsb]{202, 0.75, 1.0}{\strut .}} They are now playing in the 4th league.... & \textbf{L9} & 
                Polonia was relegated from the country's top flight in 2013 {\setlength{\fboxsep}{0pt}\colorbox[Hsb]{202, 0.15, 1.0}{\strut because}} of {\setlength{\fboxsep}{0pt}\colorbox[Hsb]{202, 0.35, 1.0}{\strut their}} {\setlength{\fboxsep}{0pt}\colorbox[Hsb]{202, 0.75, 1.0}{\strut disastrous}} financial {\setlength{\fboxsep}{0pt}\colorbox[Hsb]{202, 0.5, 1.0}{\strut situation}}{\setlength{\fboxsep}{0pt}\colorbox[Hsb]{202, 0.25, 1.0}{\strut .}} They are now playing in the 4th league....
            \end{tabular}     
             \begin{tabular}{p{0.25cm} p{6.5cm} p{0.25cm} p{6.5cm}}
                \textbf{L1} & Polonia was {\setlength{\fboxsep}{0pt}\colorbox[Hsb]{202, 0.35, 1.0}{\strut relegated}} from the country's top flight in {\setlength{\fboxsep}{0pt}\colorbox[Hsb]{202, 0.5, 1.0}{\strut 2013}} {\setlength{\fboxsep}{0pt}\colorbox[Hsb]{202, 0.75, 1.0}{\strut because}} of {\setlength{\fboxsep}{0pt}\colorbox[Hsb]{202, 0.16, 1.0}{\strut their}} {\setlength{\fboxsep}{0pt}\colorbox[Hsb]{202, 0.25, 1.0}{\strut disastrous}} financial situation. They are now playing in the 4th league.... & \textbf{L10} &
                Polonia was relegated from the country's top flight in 2013 
                {\setlength{\fboxsep}{0pt}\colorbox[Hsb]{202, 0.35, 1.0}{\strut because}} of {\setlength{\fboxsep}{0pt}\colorbox[Hsb]{202, 0.5, 1.0}{\strut their}} {\setlength{\fboxsep}{0pt}\colorbox[Hsb]{202, 0.75, 1.0}{\strut disastrous}} financial {\setlength{\fboxsep}{0pt}\colorbox[Hsb]{202, 0.25, 1.0}{\strut situation}}{\setlength{\fboxsep}{0pt}\colorbox[Hsb]{202, 0.15, 1.0}{\strut .}} They are now playing in the 4th league....
            \end{tabular}  
             \begin{tabular}{p{0.25cm} p{6.5cm} p{0.25cm} p{6.5cm}}
                \textbf{L2} & Polonia was relegated from the country's top flight in 2013 {\setlength{\fboxsep}{0pt}\colorbox[Hsb]{202, 0.75, 1.0}{\strut because}} of {\setlength{\fboxsep}{0pt}\colorbox[Hsb]{202, 0.25, 1.0}{\strut their}} {\setlength{\fboxsep}{0pt}\colorbox[Hsb]{202, 0.5, 1.0}{\strut disastrous}} financial {\setlength{\fboxsep}{0pt}\colorbox[Hsb]{202, 0.15, 1.0}{\strut situation}}. {\setlength{\fboxsep}{0pt}\colorbox[Hsb]{202, 0.35, 1.0}{\strut They}} are now playing in the 4th league.... & \textbf{L11} &  
                Polonia was relegated from the country's top flight in 2013 {\setlength{\fboxsep}{0pt}\colorbox[Hsb]{202, 0.25, 1.0}{\strut because}} of {\setlength{\fboxsep}{0pt}\colorbox[Hsb]{202, 0.5, 1.0}{\strut their}} {\setlength{\fboxsep}{0pt}\colorbox[Hsb]{202, 0.15, 1.0}{\strut disastrous}} financial {\setlength{\fboxsep}{0pt}\colorbox[Hsb]{202, 0.75, 1.0}{\strut situation}}{\setlength{\fboxsep}{0pt}\colorbox[Hsb]{202, 0.35, 1.0}{\strut .}} They are now playing in the 4th league....
            \end{tabular}
        \end{tabular}  
    \caption{Heatmap visualisation of the $I_l$ distribution over BERT's first and last 3 layers, for a sample from SQuAD. The initial layers focus on question specific words and latter focus on supporting words that lead to answer}.
    \label{tab:qual_eg}
\end{table*}

\subsection{Probing layers: QA functionality}
Based on the defined layers' functionality $I_l$, we try to identify which layers focus more on the question, the context around the answer, etc. We segregate the passage words into three categories: \textit{answer words, supporting words, and query words}, where supporting words are the words surrounding the answer within a window size of 5. Query words are the question words which appear in the passage. We take the top-5 words marked as important in $I_l$ for any layer $l$ and compute how many words from each of the above-defined categories appear in the top-5 words (results in Tables \ref{tab:semantic_stats_squad} and \ref{tab:semantic_stats_duo}).
We observe similar overall trends for both SQuAD and DuoRC. From Column 3, it is evident that the model first tries to identify the part of the passage where the question words are present. As it gets more confident about the answer (Column 2), the question words' importance decreases. From Col. 4, we infer that the layers' contextual role increases from the initial to the final layers.

\noindent \textbf{Qualitative Example:} We present a visualization of the top-5 words of the first and last three layers (with respect to $I_l$) in Table \ref{tab:qual_eg} for a sample from SQuAD. We see that all six layers give a high score to the answer span itself (`disastrous', `situation'). Further, we see that the initial layers 0,1 and 2 are also trying to connect the passage and the query (`relegated', `because',  `Polonia' get high importance scores). Hence, in this example, we see that the initial layers incorporate interaction between the query and passage. In contrast, the last layers focus on enhancing and verifying the model's prediction. 

\subsection{Visualizing Word Representations}
\begin{figure}[h]
    \centering
    \begin{subfigure}{0.41\textwidth}
        \centering
        \includegraphics[width=\textwidth]{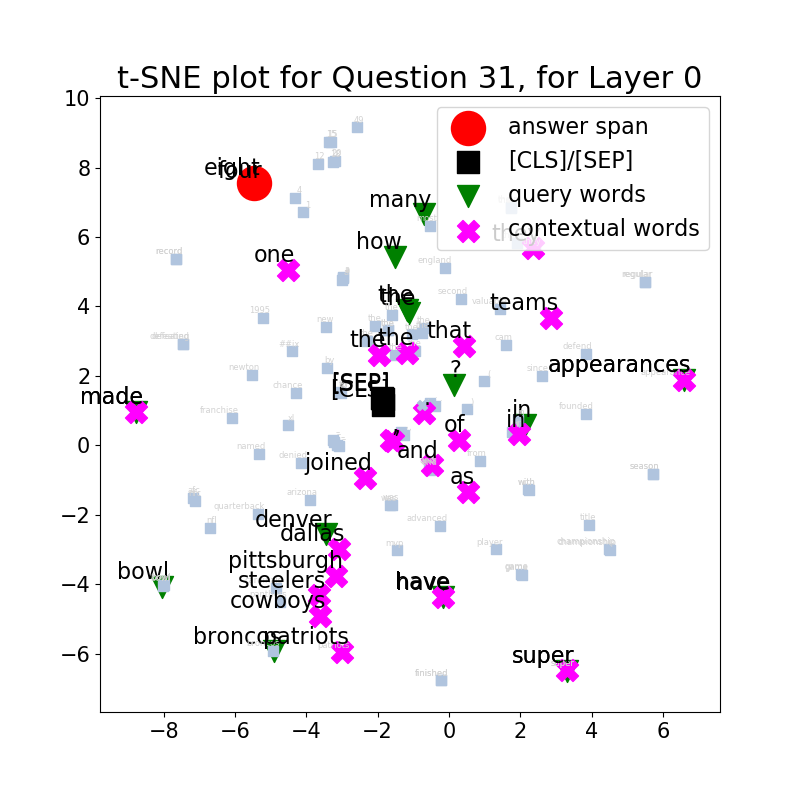} 
    \end{subfigure}
     \begin{subfigure}{0.41\textwidth}
        \centering
        \includegraphics[width=\textwidth]{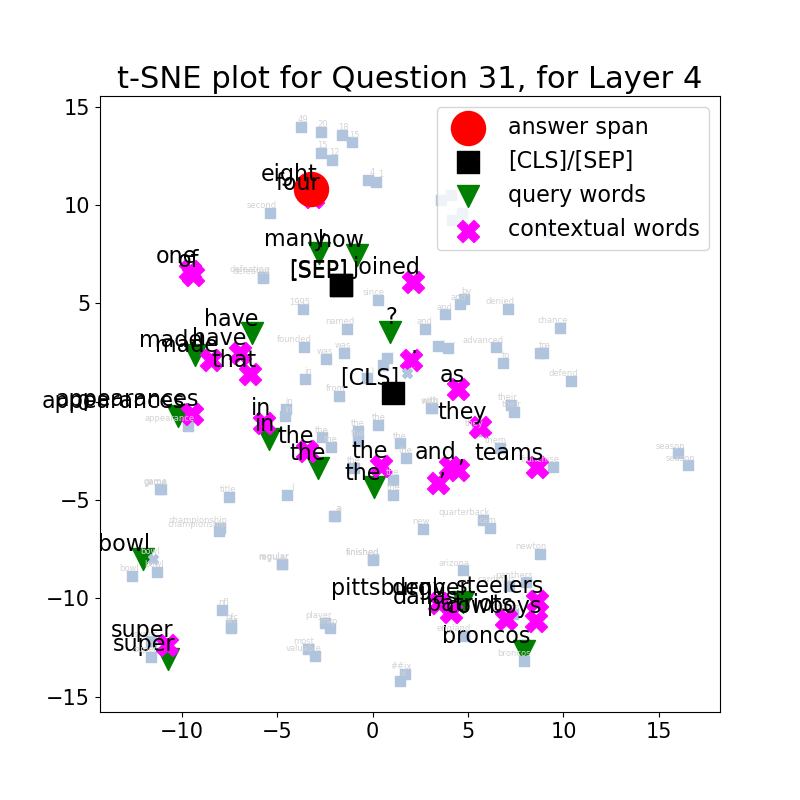} 
    \end{subfigure} 
    \caption{t-SNE plots - word embeddings of layers 0, 4 for the example in Table \ref{tab:eg_squad}. For layer 0 similar words (e.g., team names, stop words) are close to each other. For intermediate layers like Layer 4, all the contextual, answer and question words intermingle.}
    \label{fig:tsne1}
\end{figure}

\begin{figure}[h] 
    \centering
     \begin{subfigure}{0.41\textwidth}
        \centering
        \includegraphics[width=\textwidth]{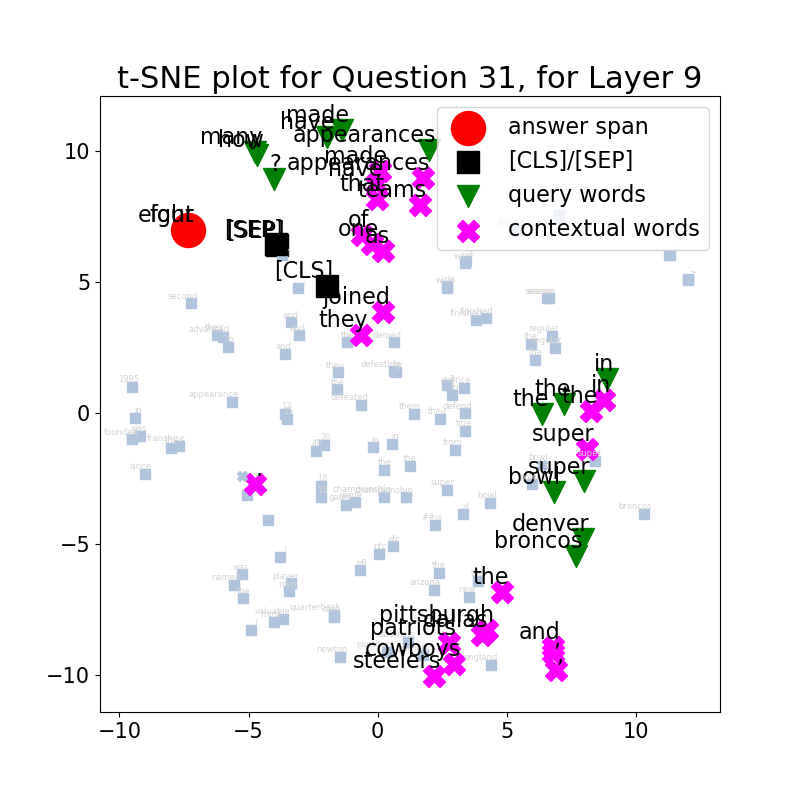} 
    \end{subfigure}
     \begin{subfigure}{0.41\textwidth}
        \centering
        \includegraphics[width=\textwidth]{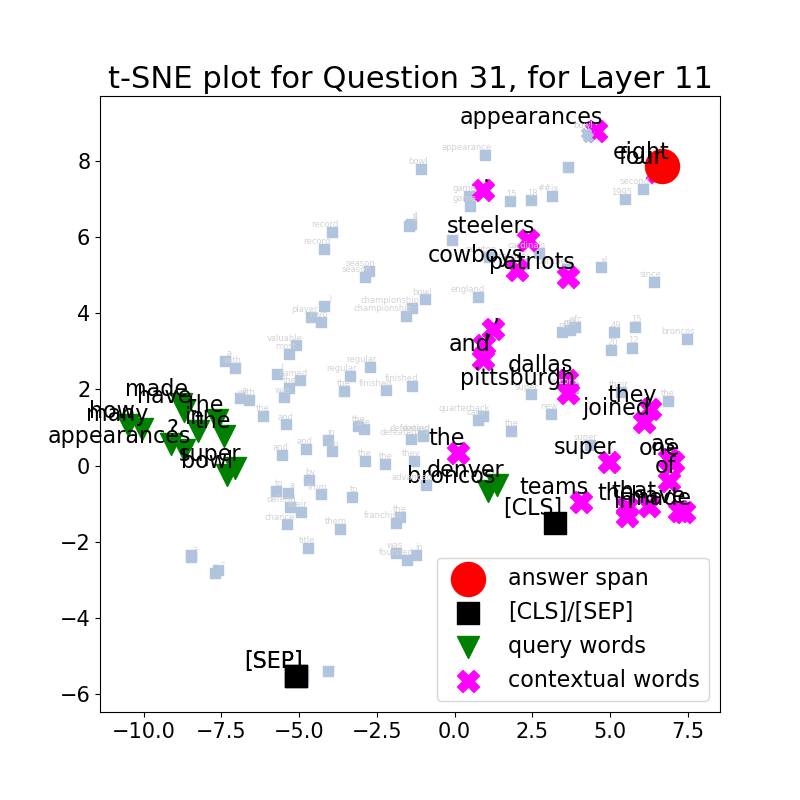} 
    \end{subfigure} 
    \caption{t-SNE plots- word embeddings of layers 9, 11 for the example in Table \ref{tab:eg_squad}. In layers 9-11, the answer \textit{eight} segregates from other words. However, numerical entity \textit{four}, is very close to the answer.}
    \label{fig:tsne2}
\end{figure}

We now qualitatively analyze the word representations of each layer. We visualize the t-SNE plot for one such {passage, question,answer} triplet from SQuAD (refer Table \ref{tab:eg_squad}) in  Figures \ref{fig:tsne1}, \ref{fig:tsne2}. We visualize the answer, supporting words, query words, and special tokens. Note that we have grayed out the other words in the passage. In initial layers (such as layer 0), we observe that similar words such as stop-words, team names, numbers \{eight, four\}, etc., are close to each other. In Layer 4, the passage, question, and answer come closer to each other. By layer 9, we see that the answer words are segregated from the rest of the words, even though the passage word `four', which is of the same type as the answer `eight' (number), is still close to `eight'. We see more interesting observations yet here: (i) in later layers, the question words separate from the answer and the supporting words, (ii) Across all 12 layers, embeddings for \textit{four, eight} remain very close together, which could have easily led to the model making a wrong prediction. However, the model still predicts the answer `eight' correctly. We were not able to identify the layer where the distinction between the two confusing answers occurs.  
\begin{table}
    \centering
    \begin{tabular}{|p{7cm}|}
        \hline 
         \textbf{Passage:} \textcolor{gray}{the panthers finished the regular season with a 15 – 1 record, ... the broncos ... finished the regular season with a 12 – 4 record.} \textcolor{purple}{They joined the patriots , dallas cowboys , and pittsburgh steelers as one of teams that have made \textcolor{blue}{\textbf{eight}} appearances in the super bowl} .\\
         \hline
         \textbf{Question:} \textcolor{applegreen}{\textbf{How many appearances have the Broncos made in the super bowl?}} \\
         \hline
    \end{tabular}
    \caption{Sample from the dev-split of SQuAD. \textbf{\textcolor{blue}{Blue}} shows the answer, \textcolor{purple}{purple} shows the contextual passage words and \textbf{\textcolor{applegreen}{green}} shows the query}.
    \label{tab:eg_squad}
    \vspace{-6mm}
\end{table}

\noindent\textbf{Quantifier questions:} For a detailed analysis of quantifier questions like \textit{how many, how much} that could have many confusing answers (i.e., numerical words) in the passage, we perform further analysis. Based on our layer-level functionality $I_l$, we compute the number of words that are numerical quantities in the top-5 words, and the entire passage, and compute their ratio. This represents the ratio of confusing words that are marked as important by each layer. There are 799 and 310 such questions in SQuAD and DuoRC, respectively.

\noindent Interestingly, we observe that this ratio \textit{increases} as we go higher up (\underline{SQuAD:} L0 - 5.6\%, L10 - 17.7\%, L11 - 15.5\%, \underline{DuoRC:} L0 - 12.9\%, L10 - 21.6\%, L11 - 22.6\%). For the example in Table \ref{tab:eg_squad}, we observed that in its later layers, BERT gives high importance to the words `eight', `four', and `second' (numerical quantities), even though the latter are not related or necessary to answer the question. This shows that BERT, in its later layers, distributes its focus over confusing words. However, it \textit{still} manages to predict the correct answer for such questions (87.35\% EM for such questions for SQuAD, and 53.5\% in DuoRC); BERT also has high confidence in predicting the answer for such questions (86.5\% vs 80.4\% for quantifier questions with more than one numerical entity in the passage vs non-quantifier questions in SQuAD, 95.2\% vs 87.2\% in DuoRC). 
This behavior is very different from the assumed roles a layer might take to answer the question, as it is expected that such words were considered in the initial rather than final layers. This shows the complexity of BERT and the difficulty of interpreting it for the RCQA task.
\section{Conclusion} 
In this work, we highlight that the lack of pre-defined roles for layers adds to the difficulty of interpreting highly complex BERT-based models. We first define each layer's functionality using Integrated Gradients. We present results and analysis to show that BERT is learning some form of passage-query interaction in its initial layers before arriving at the answer. We found the following observations interesting and with a potential to be probed further: (i) why do the question word representations move away from contextual and answer representation in later layers? (ii) If the focus on confusing words increases from the initial to later layers, how does BERT still have a high accuracy? We hope that this work will help the research community interpret BERT for other complex tasks and explore the above open-ended questions. 

\section*{Acknowledgements}
We thank the Department of Computer Science and Engineering, IIT Madras and the Robert Bosch Center for  Data  Science  and  Artificial  Intelligence, IIT Madras (RBC-DSAI) for providing us compute resources. We thank Google for supporting Preksha Nema contribution through the Google Ph.D. Fellowship programme. We also thank the anonymous reviewers for their valuable and constructive suggestions. 

\bibliographystyle{acl_natbib}
\bibliography{why_bert_difficult}

\appendix
\begin{table*}[h]
\begin{minipage}{\textwidth}
\centering
\resizebox{0.8\textwidth}{!}{%
\begin{tabular}{|c|c|c|c|c|c|c|c|}
\hline
\textbf{Layer Name} & \textbf{\begin{tabular}[c]{@{}c@{}}\% common / proper /\\ cardinal nouns\end{tabular}} & \textbf{\% verbs} & \textbf{\% stop words} & \textbf{\% adverbs} & \textbf{\% adjectives} & \textbf{\% punct marks} & \textbf{\begin{tabular}[c]{@{}c@{}}\% words in \\ answer span\end{tabular}} \\ \hline
Layer 0 & 49.57 & 12.92 & 12.63 & 2.73 & 11.63 & 11.41 & 26.99 \\ \hline
Layer 1 & 53.65 & 13.81 & 10.71 & 3.13 & 11.44 & 8.27 & 26.09 \\ \hline
Layer 2 & 52.16 & 14.19 & 13.71 & 3.24 & 12.52 & 5.25 & 29.9 \\ \hline
Layer 3 & 49.63 & 12.98 & 16.27 & 2.76 & 10.97 & 8.52 & 30.44 \\ \hline
Layer 4 & 47.99 & 12.32 & 19.93 & 2.87 & 10.58 & 7.29 & 30.06 \\ \hline
Layer 5 & 46.97 & 12.34 & 19.35 & 2.73 & 9.56 & 10.29 & 30.75 \\ \hline
Layer 6 & 49.61 & 12.13 & 17.38 & 2.51 & 9.74 & 9.75 & 31.25 \\ \hline
Layer 7 & 50.43 & 11.31 & 16.23 & 2.61 & 9.87 & 10.85 & 32.37 \\ \hline
Layer 8 & 54.16 & 11.59 & 14.59 & 2.58 & 11.27 & 6.94 & 30.78 \\ \hline
Layer 9 & 53.09 & 10.11 & 12.98 & 2.42 & 11.01 & 11.82 & 34.58 \\ \hline
Layer 10 & 57.8 & 8.67 & 12.2 & 2.11 & 10.93 & 9.64 & 34.31 \\ \hline
Layer 11 & 54.58 & 8.77 & 14.57 & 2.31 & 10.43 & 10.63 & 34.63 \\ \hline
\end{tabular}%
}
\caption{Part-of-Speech statistics of top-5 words - SQuAD}
\label{tab:postag_stats_squad}
\end{minipage}
\begin{minipage}{\textwidth}
\centering
\resizebox{0.8\textwidth}{!}{%
\begin{tabular}{|c|c|c|c|c|c|c|c|}
\hline
\textbf{Layer Name} & \textbf{\begin{tabular}[c]{@{}c@{}}\% common / proper /\\ cardinal nouns\end{tabular}} & \textbf{\% verbs} & \textbf{\% stop words} & \textbf{\% adverbs} & \textbf{\% adjectives} & \textbf{\% punct marks} & \textbf{\begin{tabular}[c]{@{}c@{}}\% words in \\ answer span\end{tabular}} \\ \hline
Layer 0 & 55.81 & 12.63 & 9.5 & 1.97 & 9.56 & 10.87 & 35.14 \\ \hline
Layer 1 & 58.1 & 13.21 & 8.41 & 2.16 & 10.03 & 8.6 & 37.29 \\ \hline
Layer 2 & 59.42 & 13.9 & 8.67 & 2.22 & 10.54 & 5.61 & 38.30 \\ \hline
Layer 3 & 55.03 & 13.61 & 11.55 & 2.15 & 9.54 & 8.78 & 34.37 \\ \hline
Layer 4 & 54.43 & 13.91 & 12.63 & 1.97 & 9.14 & 8.26 & 33.93 \\ \hline
Layer 5 & 51.97 & 13.09 & 12.58 & 1.82 & 8.04 & 12.79 & 36.32 \\ \hline
Layer 6 & 54.88 & 12.35 & 9.84 & 1.77 & 8.45 & 12.88 & 35.34 \\ \hline
Layer 7 & 60.12 & 10.02 & 9.34 & 1.8 & 9.07 & 9.94 & 41.20 \\ \hline
Layer 8 & 60.81 & 8.56 & 7.64 & 1.84 & 9.2 & 12.33 & 40.38 \\ \hline
Layer 9 & 60.96 & 8.84 & 8.2 & 1.84 & 9.24 & 11.33 & 41.25 \\ \hline
Layer 10 & 57.43 & 8.42 & 10.57 & 1.81 & 9.05 & 13.24 & 43.93 \\ \hline
Layer 11 & 60.46 & 9.07 & 11.06 & 1.97 & 9.39 & 8.65 & 44.37 \\ \hline
\end{tabular}%
}
\caption{Part-of-Speech statistics of top-5 words - DuoRC}
\label{tab:postag_stats_duorc}
\end{minipage}
\end{table*}
\section{Probing layers: POS Tags} \label{sec:appendix}
Based on the layers' functionality $I_l$, we analyze the top-5 important words in each layer on the basis of POS tags. The results can be found in Tables \ref{tab:postag_stats_squad} and \ref{tab:postag_stats_duorc}. We note that all 12 layers are majorly focused on entity based words (common nouns, proper nouns and numerical entities). Surprisingly, all layers give approximately 10\% of their importance to punctuation marks and stopwords each, the same level of importance that is given to verbs and adjectives. It is worth noting that on average, answer spans in SQuAD on 82.04\% entites, and answer spans in DuoRC are 79.78\% entities.

\end{document}


\appendix

\section{Appendices}
\label{sec:appendix}
Appendices are material that can be read, and include lemmas, formulas, proofs, and tables that are not critical to the reading and understanding of the paper. 
Appendices should be \textbf{uploaded as supplementary material} when submitting the paper for review.
Upon acceptance, the appendices come after the references, as shown here.

\paragraph{\LaTeX-specific details:}
Use {\small\verb|\appendix|} before any appendix section to switch the section numbering over to letters.

\subsection{Models used}

We describe here in detail the four models that we analyse in our work and explain how they are special instances of the generic model framework.\\

\noindent \textbf{DCN}: The word embedding layer in DCN encodes passage and question words using pre-trained word-level embeddings. The contextual layer uses a uni-directional LSTM. The cross-interaction layer captures interactions between passage and question words using a series of linear transformations of the pairwise word-to-word similarity matrix. DCN does not have a modeling layer. The final output layer contains an LSTM-based sequential model to predict start and end positions using Highway Maxout Networks iteratively.

\noindent \textbf{BiDAF}: The word-embedding layer in BiDAF  contains a Highway network which embeds passage and question words using both word and character-level information. The contextual layer contains a bidirectional LSTM. The cross interaction layer fuses passage and question word representations to yield question-aware passage representation and passage-aware question representation. The modeling layer further refines the passage word representations using two layers of Bi-directional LSTM. The output layer is a simple linear-layer which predicts the answer start and answer end indices.

\noindent \textbf{QANET}: QANET's architecture is similar to that BiDAF. The primary difference is that an ``Encoding Block'' replaces all the Bi-directional LSTMs across all layers in BiDAF. Here, the Encoding Block is a stack of a variable number of convolution layers, a self-attention layer, and a feed-forward layer. 

\noindent \textbf{BERT}: BERT is a deep bi-directional transformer-based model designed to pre-train word representations from unlabeled text.  
It has 12 Transformer blocks (layers) wherein each block(layer) has a multi-head self-attention and a feed-forward neural network. Each layer allows for interactions between query and passage words as well as between passage words. Thus we can think of each layer as facilitating both self interactions and cross interactions. At each layer a refined representation of every passage word is computed based using a transformer block. The output layer then takes the final refined representation for a passage word as input and predicts whether it is a start index or end index or neither.

\subsection{Layer Level Functionality}
\subsection{JSD heatmaps for BiDAF, DCN, QANet}
\begin{figure*}[h]
    \centering
    \begin{subfigure}{0.24\textwidth}
    \centering
    \includegraphics[width=\textwidth]{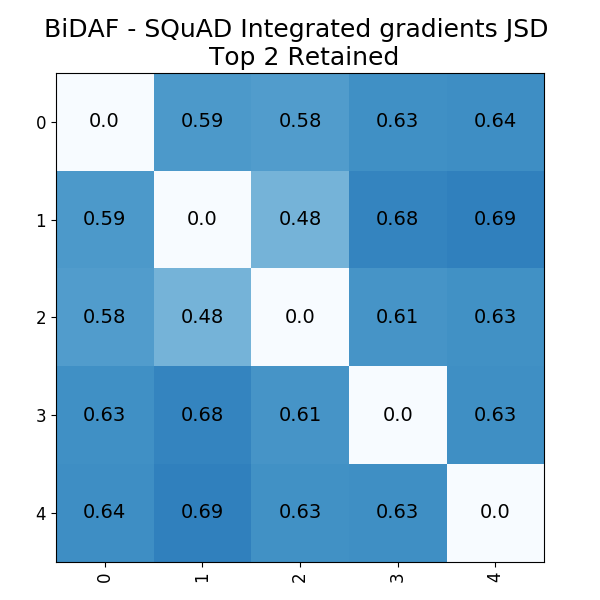}
    \caption{}
    \label{bidaf_jsd1}
    \end{subfigure}
    \begin{subfigure}{0.24\textwidth}
    \centering
    \includegraphics[width=\textwidth]{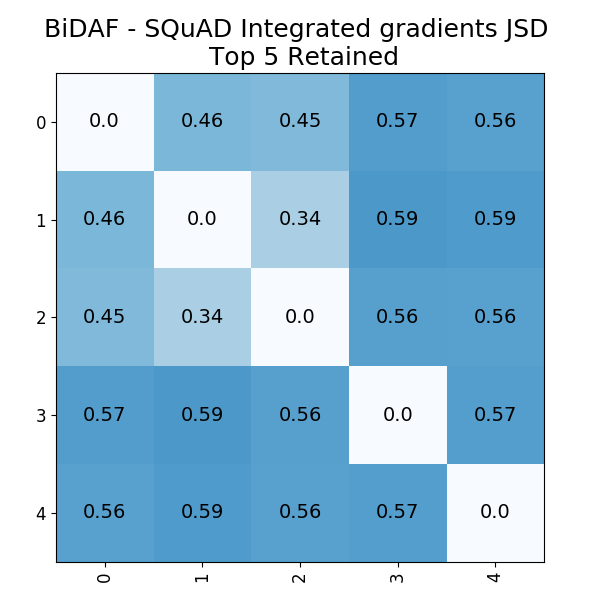}
    \caption{}
    \label{bidaf_jsd2}
    \end{subfigure}
    \begin{subfigure}{0.24\textwidth}
    \centering
    \includegraphics[width=\textwidth]{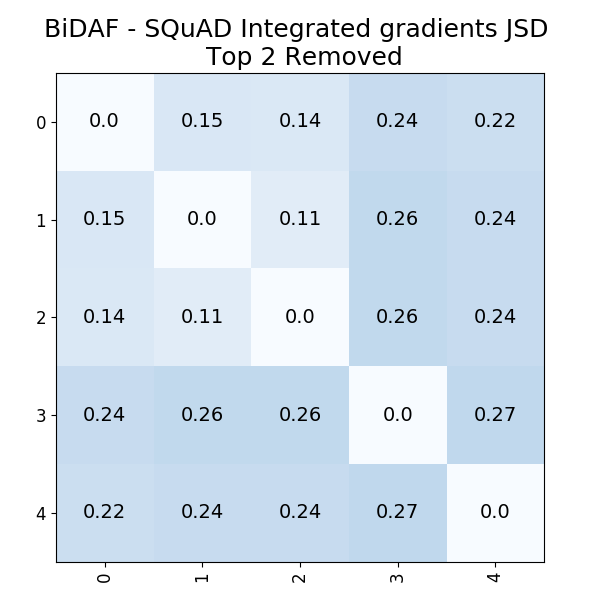}
    \caption{}
    \label{bidaf_jsd3}
    \end{subfigure}
    \begin{subfigure}{0.24\textwidth}
    \centering
    \includegraphics[width=\textwidth]{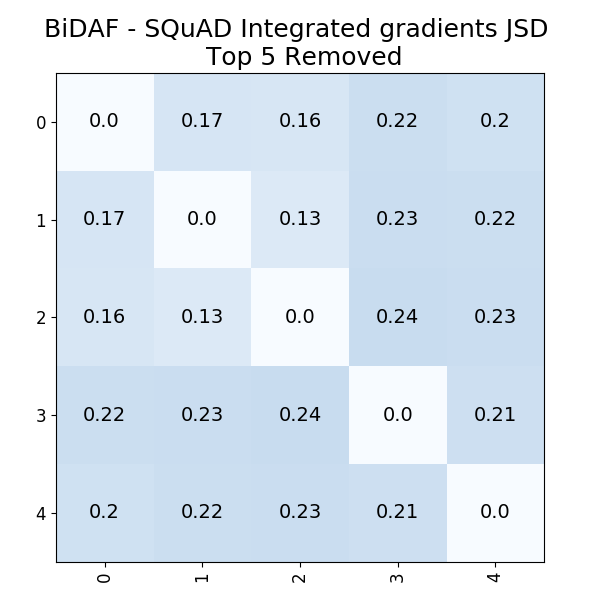}
    \caption{}
    \label{bidaf_jsd4}
    \end{subfigure}\\
    \begin{subfigure}{0.24\textwidth}
    \centering
    \includegraphics[width=\textwidth]{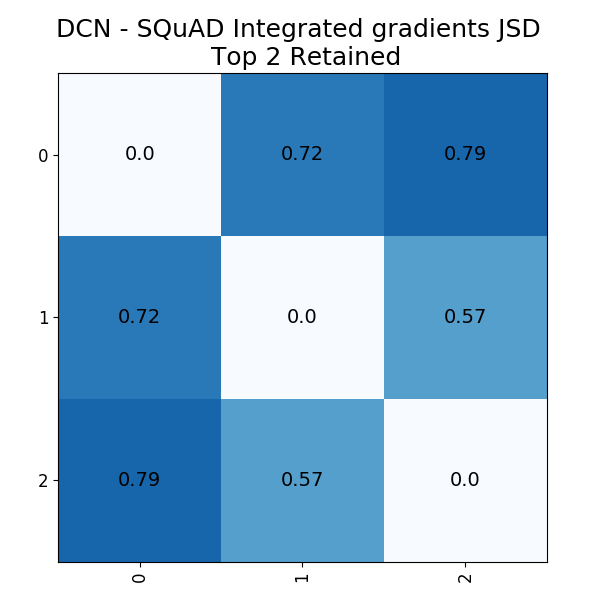}
    \caption{}
    \label{dcn_jsd1}
    \end{subfigure}
    \begin{subfigure}{0.24\textwidth}
    \centering
    \includegraphics[width=\textwidth]{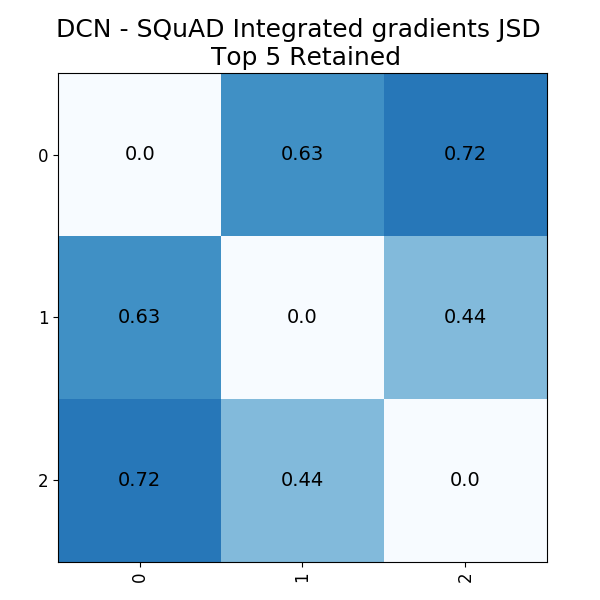}
    \caption{}
    \label{dcn_jsd2}
    \end{subfigure}
    \begin{subfigure}{0.24\textwidth}
    \centering
    \includegraphics[width=\textwidth]{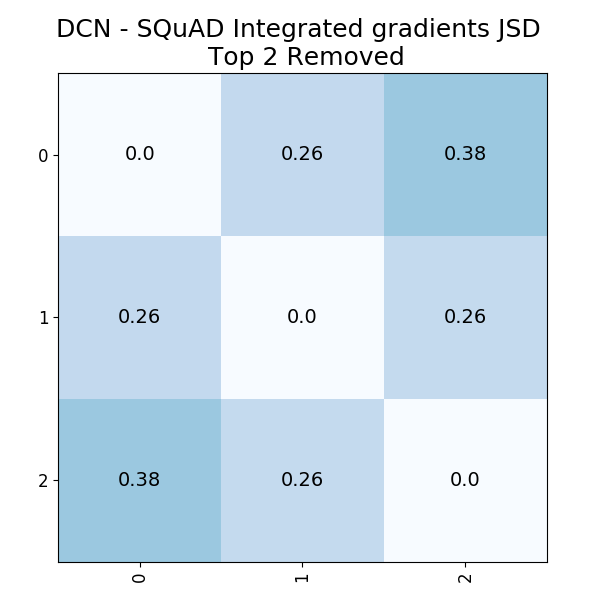}
    \caption{}
    \label{dcn_jsd3}
    \end{subfigure}
    \begin{subfigure}{0.24\textwidth}
    \centering
    \includegraphics[width=\textwidth]{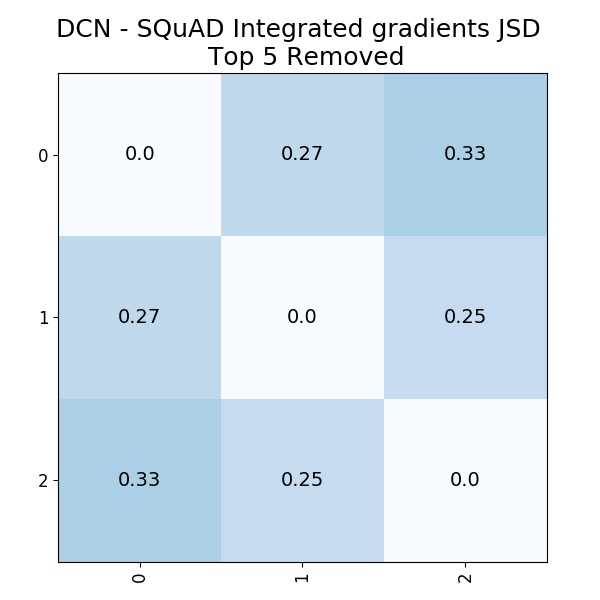}
    \caption{}
    \label{dcn_jsd4}
    \end{subfigure}\\    
    \begin{subfigure}{0.24\textwidth}
    \centering
    \includegraphics[width=\textwidth]{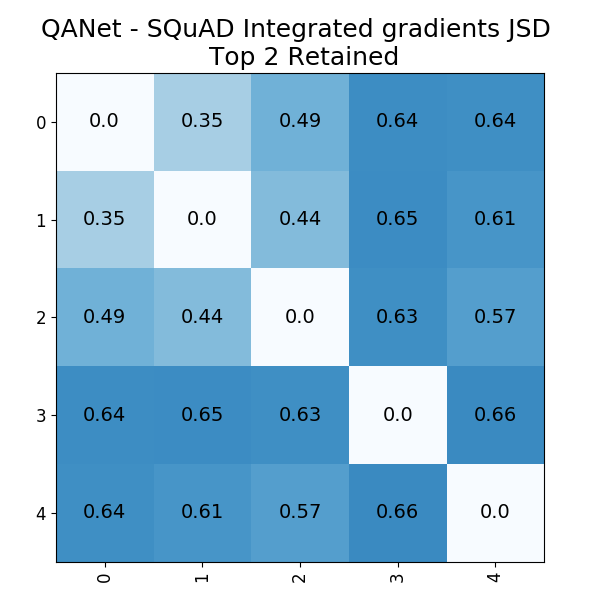}
    \caption{}
    \label{qanet_jsd1}
    \end{subfigure}
    \begin{subfigure}{0.24\textwidth}
    \centering
    \includegraphics[width=\textwidth]{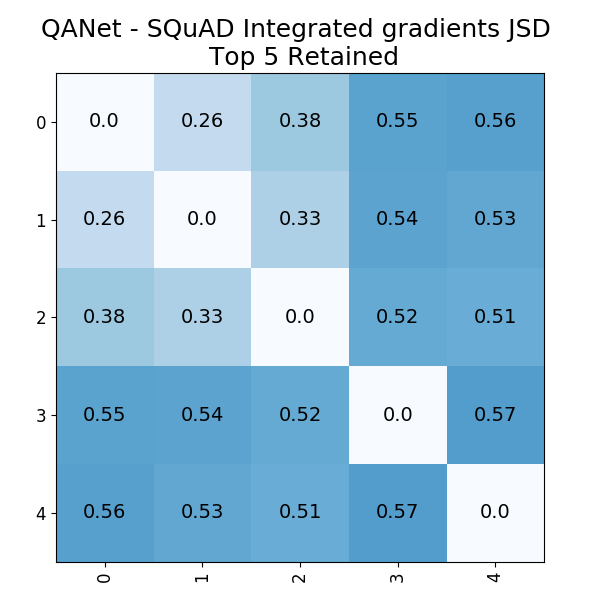}
    \caption{}
    \label{qanet_jsd2}
    \end{subfigure}
    \begin{subfigure}{0.24\textwidth}
    \centering
    \includegraphics[width=\textwidth]{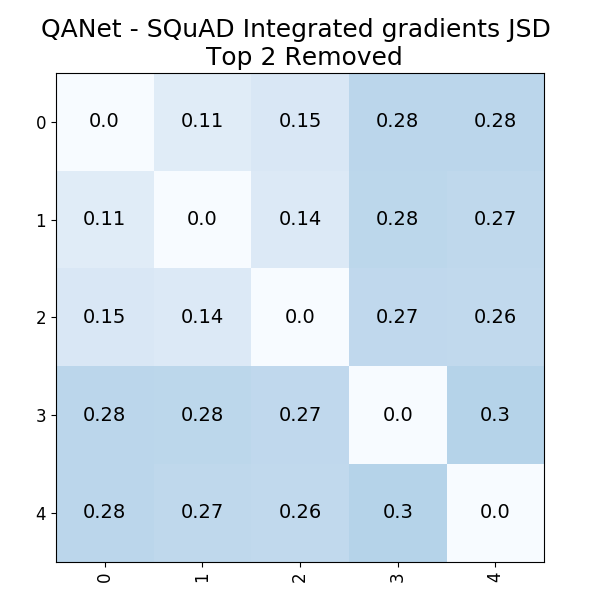}
    \caption{}
    \label{qanet_jsd3}
    \end{subfigure}
    \begin{subfigure}{0.24\textwidth}
    \centering
    \includegraphics[width=\textwidth]{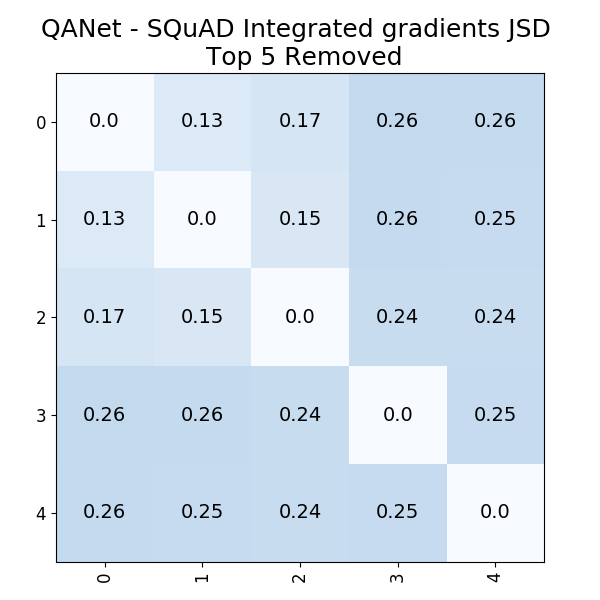}
    \caption{}
    \label{qanet_jsd4}
    \end{subfigure}\\      
    \caption{JSD of importance score distributions with top-k items removed/retained, averaged across 500 datapoints of SQuAD for BiDAF, DCN and QANet (top to bottom rows)}
    \label{fig:others_jsd_rem_keep}
\end{figure*}

Remaining JSD heatmaps for BiDAF,DCN and QANet, can be found in Figure \ref{fig:others_jsd_rem_keep}.

\subsection{Analysis for more question types}
Remaining JSD heatmaps for BERT split by question types, can be found in Figure \ref{fig:bert_jsd_qtype_more}.
\begin{figure*}[h]
    \centering
    \begin{subfigure}{0.32\textwidth}
    \centering
    \includegraphics[width=\textwidth]{jsd_qtype/when_bert_squad_js_keep2_avg_500.png}
    \caption{}
    \label{bert_jsd_when}
    \end{subfigure}
    \begin{subfigure}{0.32\textwidth}
    \centering
    \includegraphics[width=\textwidth]{jsd_qtype/where_bert_squad_js_keep2_avg_500.png}
    \caption{}
    \label{bert_jsd_where}
    \end{subfigure}
    \begin{subfigure}{0.32\textwidth}
    \centering
    \includegraphics[width=\textwidth]{jsd_qtype/which_bert_squad_js_keep2_avg_500.png}
    \caption{}
    \label{bert_jsd_which}
    \end{subfigure}
    \caption{JSD of importance score distributions, averaged across 500 datapoints of SQuAD for BERT, split by question types}
    \label{fig:bert_jsd_qtype_more}
\end{figure*}